\documentclass{article}
\usepackage{amssymb}

\usepackage[final]{corl_2025} 

\usepackage[T1]{fontenc}
\usepackage{cite}
\usepackage{caption}
\usepackage{graphicx}
\usepackage{subcaption}
\usepackage{xcolor}
\usepackage{color}
\usepackage{amssymb}
\usepackage{float}
\usepackage{amsmath}
\usepackage{url}
\usepackage{booktabs}
\usepackage{tabularx}

\title{KoopMotion: Learning Almost Divergence-Free Koopman Flow Fields
for Motion Planning}

\author{
  Alice Kate Li, Thales C. Silva, Victoria Edwards, Vijay Kumar, M. Ani Hsieh\\
  GRASP Laboratory\\
  University of Pennsylvania, 
  United States\\
  \texttt{\{alicekl, scthales, vmedw, kumar, mya\}@seas.upenn.edu} \\
}

\begin{document}
\maketitle

\begin{abstract}
In this work, we propose a novel flow field-based motion planning method that
drives a robot from any initial state to a desired reference trajectory such that it converges to the trajectory's end point. 
Despite demonstrated efficacy in using Koopman operator theory for modeling dynamical systems, Koopman does not inherently enforce convergence to desired trajectories nor to specified goals---a requirement when learning from demonstrations (LfD). 
We present KoopMotion which represents motion flow fields as dynamical systems, parameterized by Koopman Operators to mimic desired trajectories, and leverages the divergence properties of the learnt flow fields to obtain smooth motion fields that converge to a desired reference trajectory when a robot is placed away from the desired trajectory, and tracks the trajectory until the end point.
To demonstrate the effectiveness of our approach, we show evaluations of KoopMotion on the LASA human handwriting dataset and a 3D manipulator end-effector trajectory dataset, including spectral analysis.
We also perform experiments on a physical robot, verifying KoopMotion on a miniature autonomous surface vehicle operating in a non-static fluid flow environment.
Our approach is highly sample efficient in both space and time, requiring only 3\% of the LASA dataset to generate dense motion plans.
Additionally, KoopMotion provides a significant improvement over baselines when comparing metrics that measure spatial and temporal dynamics modeling efficacy. Code at: \href{https://alicekl.github.io/koop-motion/}{\color{blue}{https://alicekl.github.io/koop-motion}}.


\end{abstract}



\section{Introduction}
Learning from demonstrations (LfD) is an imitation learning paradigm in which a robot or agent learns how to complete a task, \textit{e.g.} a particular motion, by mimicking behaviors from given demonstrations.  Learning from demonstrations is particularly useful for programming complex actions especially when desired actions cannot easily be defined as an optimization problem.
As a result, there is a rapidly growing interest in developing LfD methods for robots to learn increasingly complex tasks.
Such methods can fall under one of three categories: learning policies from demonstrations \citep{khansari2011learning}, learning cost or rewards from demonstrations \citep{dragan2015movement, dvijotham2010inverse}, and learning plans from demonstrations \citep{konidaris2012robot}.
In this paper, we focus on learning policies, by learning  dynamical systems from demonstrations. 
This dynamical system can be used to generate reference trajectories for a robot during motion planning.

To date, the most commonly used approach for learning dynamical system representations for motion planning has been Gaussian Mixture Models (GMM) \citep{khansari2011learning,  ravichandar2017learning, ravichandar2019learning, figueroa2018physically, li2023task, sun2024se}.
Other methods adopt Neural ODEs (NODE) \citep{nawaz2024learning, kasaei2023data}, normalizing flows \citep{urain2020imitationflow}, Euclidean flows \citep{rana2020euclideanizing}, or Gaussian Process Regression \citep{schneider2010robot}.
While Koopman operator theory has been shown to model nonlinear dynamical systems effectively \citep{koopman1931hamiltonian, mezic2013analysis, pan2020physics, otto2021koopman, brunton2021modern, yeung2019learning, li2017extended, shaffer2025spectrally, lusch2018deep},
few works have used Koopman for learning dynamical systems, to be used as reference trajectories that guide motion planning \citep{bevanda2022diffeomorphically, salam2023online, li2024enkode, han2023utility}.
Instead, most Koopman-based works adopt Koopman operator theory from a more control-theoretic approach, and use their Koopman models for model-based control \citep{bruder2019modeling, abraham2017model,abraham2019active, sotiropoulos2021dynamic, selby2021learning}.
However, of the few existing works that do focus on the direct learning of dynamical systems to generate reference trajectories, they do not address the convergence to desired trajectories or goals.
Since Koopman operator theoretic approaches involve modeling the nonlinear dynamics via a spectral decomposition, one can study the stability properties of the linearized systems more readily than other approaches, such as GMMs and NODEs.
Given this, we are interested in investigating whether Koopman operator theory can be used for learning dynamical systems from demonstrations, where a successfully learnt model captures the complex structure of the demonstrations, guides the system to the desired motions, and converges to the desired goal.

Our contributions are a novel approach for learning almost divergence-free Koopman dynamical system flow fields for motion planning.
More specifically, we devise two novel loss functions that guide the system towards the desired demonstration trajectories and the desired endpoint. 
We demonstrate the efficacy of these novel loss functions on learning motion plans trained on the LASA handwriting dataset.
Our approach, KoopMotion, maps these hand-written motions to robot trajectories and can be used to keep the robot tracking trajectories under disturbances, caused by, for example, unsteady ocean flows or uneven terrain for ground vehicles or wind for aerial vehicles in agricultural fields. 
We further verify the feasibility of the learnt dynamical system by evaluating our methods on a $3D$ end-effector dataset, as well as on hardware experiments, whereby KoopMotion guides a miniature autonomous surface vehicle during motion planning.



\begin{figure}[t]
  \centering
\includegraphics[width=0.95\linewidth]{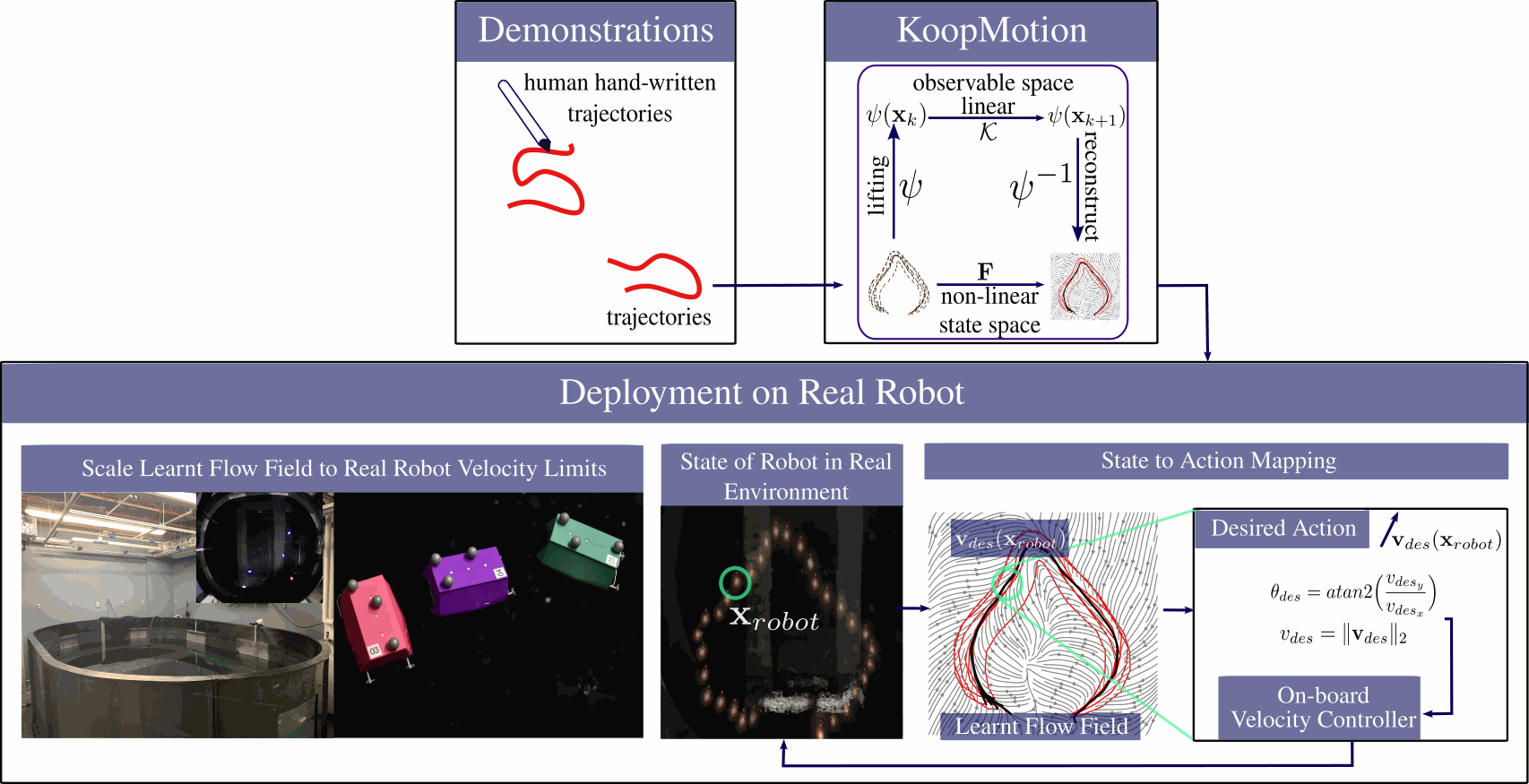}
  \caption{Overview of KoopMotion. Demonstration trajectories are acquired, representing desired nonlinear motions that a robot should follow. 
  Trajectories are inputs to the KoopMotion model, which estimates an almost divergence-free flow field for motion planning. 
  This flow field drives a robot from locations away from the desired motion towards the desired reference trajectory and towards convergence to a desired goal. 
  After scaling to the vehicle's velocity limits, learnt KoopMotion flow fields can be used for motion planning on real robots. 
  In this work, we train KoopMotion flow fields on demonstrations of human-drawn trajectories from the LASA handwriting dataset, describing the desired robot path in time. We evaluate the efficacy of using the learnt flow fields to produce reference trajectories for miniature autonomous surface vehicles navigating in a flow field tank with non-static conditions.}
  \label{fig:koopmotion_overview}
\end{figure}

\section{Problem Formulation}\label{sec:problem_formulation}

Given training data $\mathcal{D}$, consisting of desired motion demonstrations, either in the form of short horizon training pairs $(\mathbf{x}_{k}, \mathbf{x}_{k+1})$, or long horizon trajectory data of length $T$, $(\mathbf{x}_{k}, \mathbf{x}_{k+1}, \dots, \mathbf{x}_{k+T})$, where $\mathbf{x}_k$ denotes the system state at time-step $k$, learn a discrete-time nonlinear dynamical system that drives the system towards the desired reference trajectories.
We call this learnt dynamical system a \textit{motion planning flow field}, as it can be used directly to compute velocity commands for control.
The dynamical system is of the form:
\begin{equation}\label{eqn:nonlinear_dynamics}
\mathbf{x}_{k+1} = f(\mathbf{x}_k)
\end{equation}
where $f$ represents a nonlinear mapping that propagates the system forward in time. 
The system state in this work is the position 
of a mobile robot. 

\section{Methodology}
We introduce KoopMotion, our data-driven modeling approach for estimating motion planning flow fields, learnt from demonstrations.

\subsection{Koopman Operator Theory}
Due to the demonstrated effectiveness of modeling nonlinear dynamical systems from trajectory data \citep{salam2023online} and sparse flow data \citep{li2024enkode}, KoopMotion adopts a data-driven Koopman operator theoretic method to model nonlinear dynamical systems.
In short, Koopman operator theory allows us to find a linear dynamics mapping for nonlinear dynamical systems by applying an appropriate coordinate transform that lifts a system into a higher dimensional space, in which the system dynamics now evolve linearly \citep{koopman1931hamiltonian}.
Instead of defining the dynamics as in \eqref{eqn:nonlinear_dynamics}, Koopman theory involves parameterizing $f$ with lifting functions, $\Psi( \cdot )$, which are functions of the state, and the Koopman operator, $\mathcal{K}$, which is a linear operator in this lifted space. 
More specifically,
\begin{equation}\label{eqn:koopman}
\Psi(\mathbf{x}_{k+1}) = \mathcal{K}\Psi(\mathbf{x}_k).
\end{equation}
Modern Koopman theory approaches involve learning $\Psi( \cdot )$ and $\mathcal{K}$ from data \citep{ mezic2013analysis, pan2020physics, otto2021koopman, brunton2021modern, yeung2019learning, li2017extended, bevanda2022diffeomorphically, abraham2019active, bruder2019modeling, leask2021modal}, arising from work in Extended Dynamic Mode Decomposition (EDMD) \citep{williams2015data}.

\subsection{Learning Interpretable Dynamics with Fourier Features}
Similar to prior work \citep{salam2023online} and \citep{li2024enkode}, we use a set of $\nu$ Fourier features (along with the state itself) to represent the lifting functions.
This idea of predefining a set of lifting functions comes from EDMD \citep{williams2015data}, where a dictionary of different lifting functions is selected and the contribution of each lifting function required to linearize the dynamics is learnt from data. 
However, instead of having a mixture of different lifting functions, we assume that Fourier features alone are able to provide sufficient modeling capacity, whilst removing the need for an expert user to select candidate lifting functions.
The user then simply chooses the dimensionality of the lifted space. 
The lifting function that is learnt is defined as follows:
\begin{equation}\label{eqn:fourier_features}
\begin{split}
\hat \Psi(\mathbf{X}_k)= \big[\mathbf{X}_k, \text{cos}( \mathbf{w}_0^T \mathbf{X}_k + b_0), \text{cos}(\mathbf{w}_1^T \mathbf{X}_k + b_1), \\
\dots, \text{cos}(\mathbf{w}_\nu^T \mathbf{X}_k + b_\nu) \big],
\end{split}
\end{equation}%
\noindent where cosine is applied element-wise to its inputs.
Each Fourier feature has two learnable parameters $w$ and $b$. 
Intuitively, keeping the bias terms $b=0$, models with higher $w$ terms will represent flows with greater curvature in structure, or higher frequency dynamics.


\subsection{Almost Divergence-Free Flow Fields}
To shape the vector field, and encourage the KoopMotion model to learn an 
attractor along the trajectory demonstrations, we extend prior work \citep{li2024enkode}, by drawing inspiration from \citep{richter2022neural} on the divergence of vector fields.
We include an additional loss term defined by the divergence of the learnt vector field, $\mathbf{\hat{F}}$:
\begin{equation}
    \text{div}(\mathbf{\hat{F}}) = \nabla \cdot \mathbf{\hat{F}} = \sum_{i} \frac{\delta \hat{F}_i}{\delta \text{x}_i}
\end{equation} 

The learnt vector field, $\mathbf{\hat{F}}$, of the dynamical system in the original space, is estimated by computing the difference between initial conditions and forward propagated initial conditions, mapped forward in time, by one time step, via the learnt Koopman operator $\mathcal{\hat{K}}$.
Since the parameters defining $\mathbf{\hat{F}}$ arise from the composition of differentiable, learnable lifting functions, and a differentiable, learnable Koopman operator, then $\mathbf{\hat{F}}$ is also differentiable.
This differentiability allows us to compute the divergence, div($\mathbf{\hat{F}}$), via automatic differentiation.
In practice, we apply divergence constraints on a subset of the vector field domain.
We only compute the divergence of the learnt vector field at points that overlap with the training data.
The almost divergence-free flow property, defined by $\nabla \cdot\mathbf{\hat{F}}=0$ is obtained 
by using a loss term that encourages the computed divergence to be zero.
Intuitively, this means that along the desired trajectory, the flow will remain neutral, without any expansion (positive divergence) or contraction (negative divergence), minimizing any tendency of leaving the desired trajectory.



\subsection{Convergence to Goal Position}
To encourage the reference trajectories to converge to a desired goal position, we add a novel loss that is defined for parameters in the lifted space.
More specifically, the lifted system at the final time step of the trajectory, or the goal position, should be unchanged under the action of the Koopman operator:

\begin{equation}\hat{\Psi}(\mathbf{X}_{T_{final}}) =\hat{\mathcal{K}} \big( \hat{\Psi}(\mathbf{X}_{T_{final}}) \big), 
\end{equation}
where $\mathbf{X}_{T_{final}}$ represents the state of the system at the goal position. 
This should correspond to the position at the final time-step of the demonstration trajectory.

\subsection{KoopMotion Loss Functions}
We extend prior work \citep{li2024enkode}, which uses a gradient-based Koopman optimization framework to learn dynamical systems that mimic desired motions in space and time, and optimize over the two additional novel loss functions, namely a loss that encourages zero divergence of the learnt flow field around the desired trajectories, and a second loss that encourages convergence to the goal:
\begin{equation}\label{eqn:loss_function}
    \min_{{\mathcal{K},w, b}} \quad \beta_k\mathcal{L}_{Koopman} + \beta_d\mathcal{L}_{FlowDivergence} + \beta_g\mathcal{L}_{Goal} 
\end{equation}
where the losses in Equation \ref{eqn:loss_function} are defined by:
\begin{equation}\label{eqn:loss_koopman}
    \mathcal{L}_{Koopman} = \Big| \hat{\Psi}(\mathbf{X}_{k+1}) - \hat{\mathcal{K}} \big( \hat{\Psi}(\mathbf{X}_k) \big) \Big|_2,
\end{equation}
\begin{equation}\label{eqn:loss_divergence}
    \mathcal{L}_{FlowDivergence} = \Big| \sum_{i} \frac{\delta \hat{F}_i}{\delta \text{x}_i} \Big|_2,
\end{equation}
\begin{equation}\label{eqn:loss_goal}
     \mathcal{L}_{Goal} = \Big| \hat{\Psi}(\mathbf{X}_{T_{final}}) - \hat{\mathcal{K}} \big( \hat{\Psi}(\mathbf{X}_{T_{final}}) \big) \Big|_2,
\end{equation}
and where $\beta$ are weighting coefficients for each of the losses, and $\Big|\cdot\Big|_2$ denotes an mean squared error (MSE) or L2 loss.
In this work, we select $\beta_k=1$, $\beta_d=0.01$, and $\beta_g=0.01$.
An ablation study is performed in Supplementary \ref{sec:ablation_loss_term_weighting} to understand any interactions between loss terms.


\subsection{Training Data}
\textbf{2D data. } We evaluate the performance of KoopMotion on the LASA handwriting dataset from \citep{khansari2011learning}, 
which consists of desired motions for a robot that have been drawn by a human with a pen. 
In this dataset, for each motion, a human was asked to draw 7 demonstrations of a desired pattern on a tablet, by starting from different initial positions and ending at the same final point. 
The dataset consists of 30 human handwriting motions, where 26 show the motion of the pen covering motion in one direction. 
The remaining four motions include motions which start from multiple different initial conditions, and again end at the same final point. 
Learning from these handwritten motions provide motion flow fields for both a single robot or multiple robots operating in the same domain.

To train our KoopMotion model, we use planar position information only (\textit{i.e.} not velocities).
To demonstrate the sample efficiency of KoopMotion, all results shown are for sub-sampled datasets, where we take every $40$-th sample of the trajectories of $1000$ time-steps. 
This is illustrated in Fig. \ref{fig:sparse_data} of the Supplementary  \ref{sec:sparse_training_data}.

\textbf{3D data. } 
We also perform evaluations on an additional dataset of trajectories in $3D$, generated from the Robocasa simulator \citep{nasiriany2024robocasa}.
Datasets are acquired by teleoperating a 7-DOF robot moving in the simulator, and collecting the pose of the end-effector. 
Due to space limitations, results trained on $3D$ end-effector position information are shown in the Supplementary \ref{sec:modeling_3d_systems}.




\section{Results}

\begin{figure*}[t]
  \centering
\includegraphics[width=1\linewidth]{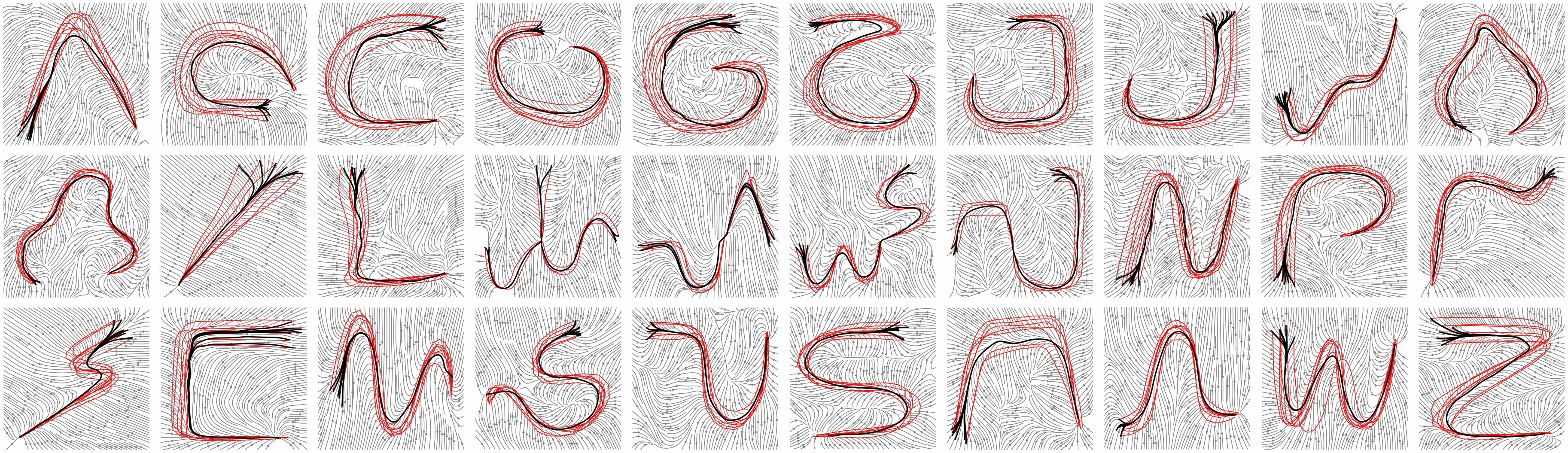}
  \caption{Qualitative performance of KoopMotion for learning motion planning flow fields for 30 nonlinear planar demonstrations from the LASA handwriting dataset. Training data, temporally sub-sampled, using less than $3\%$ of original training data (red), learnt motion plans from same initial conditions as training data (black), motion plan from all other initial conditions in the domain (gray). Eigenfunctions of some shapes are shown in Supplementary \ref{sec:supp_eigenfunctions}.}
  \label{fig:lasa_demos}
\end{figure*}

\subsection{Offline Flow Field Analysis}




\subsubsection{Goal Position Convergence Evaluations}
Learnt KoopMotion for all $30$ motions of the LASA dataset are shown in Fig. \ref{fig:lasa_demos}. 
The learnt motion plans, plotted in black, show that KoopMotion is able to capture the complex nonlinear motions in the demonstrations, even in the multi-modal examples, where demonstration trajectories start from more than one set of initial conditions.
The gray flow fields provide reference trajectories for initial conditions outside of those in the training dataset.
Upon careful inspection, qualitatively,  we observe that there are no spurious attractors in the gray flow.

\begin{figure}[t]
  \centering
   \begin{subfigure}[b]{0.4875\textwidth}
\includegraphics[width=1\linewidth]{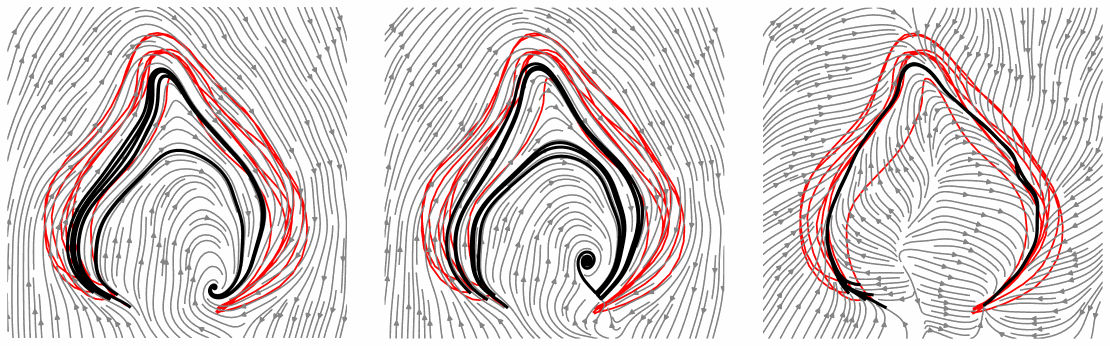}
  \caption{Effects of introducing novel losses in KoopMotion motion policies: divergence loss and goal convergence loss. \textsc{left}: $\mathcal{L}_{Koopman}$, \textsc{middle}: $\mathcal{L}_{Koopman} + \mathcal{L}_{Goal}$, and \textsc{right}: $\mathcal{L}_{Koopman}  + \mathcal{L}_{FlowDivergence} + \mathcal{L}_{Goal}$.}
  \label{fig:lasa_losses}
  \end{subfigure}
   \begin{subfigure}[b]{0.4875\textwidth}
   \includegraphics[width=1\linewidth]{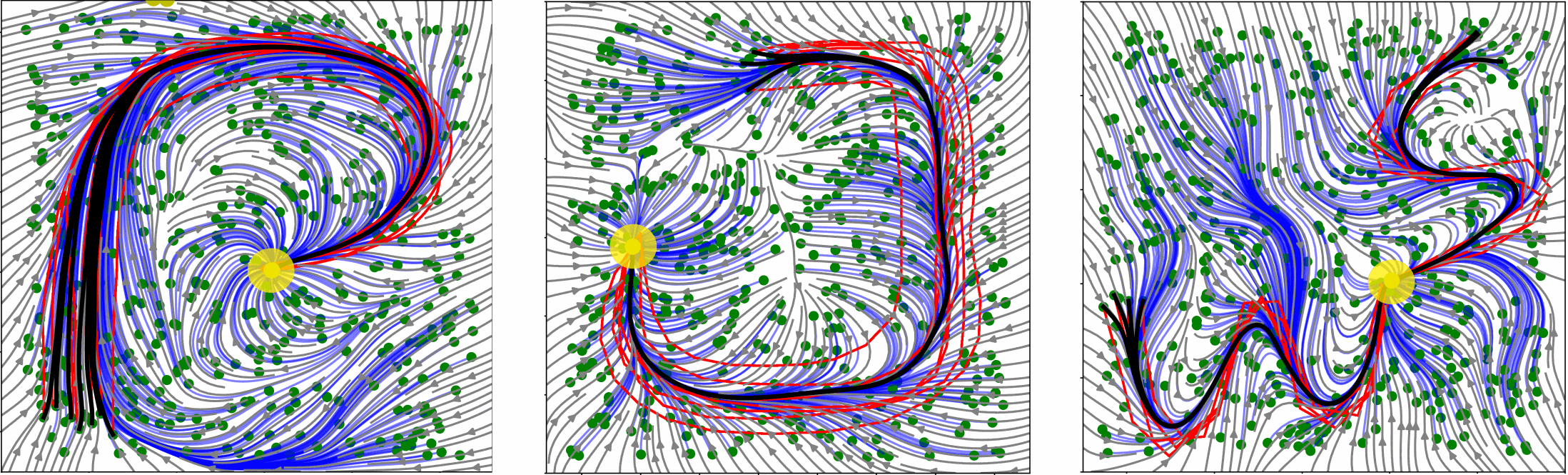}
  \caption{
  Training data (red), learnt motion plans from same initial conditions as training data (black), $500$ motion plans (blue) starting from $500$ randomly initialized initial conditions (green), with all $500$ convergence points plotted (yellow) aligning with the goal position.
  }
  \label{fig:convergence_initial_conditions}
   \end{subfigure}
     \caption{Qualitative experiments}
\end{figure}

To quantitatively evaluate the performance of KoopMotion, we take $500$ randomly selected initial conditions outside of the training dataset initial conditions, but within a square around the training set, as evaluated in \citep{khansari2011learning}, and propagate them forward in time. 
We compute the number of points that do not converge to the goal position.
We report that for all $30$ examples, all $500$ initial conditions converge to the end point of the trajectory, or hit a boundary wall. 
For these cases, we perform interference on KoopMotion to obtain the vector field for a larger domain, to ensure that those initial conditions that have hit a boundary wall would eventually converge to the goal position.
To better illustrate this qualitative evaluation, we include Fig. \ref{fig:convergence_initial_conditions}, that shows the distribution of the $500$ randomly seeded initial conditions, and how they all converge to the yellow goal position.

To highlight the effects of including the two novel losses, we include Fig. \ref{fig:lasa_losses} which shows that with only a Koopman based loss, the motion flow field almost converges to the desired goal.
With the inclusion of the goal converging loss, a spurious attractor is introduced, which is undesirable for motion planning.
However, with all three losses combined, we have both goal convergence, as well as desired flows from initial conditions outside of the training data.
Our intuition for this behavior is that the addition of $L_{Goal}$ may introduce conflicting optimization convergence criteria compared to using only $L_{Koopman}$, prohibiting the trajectory from converging to the goal.
However, adding $L_{FlowDivergence}$ regularizes the solution.

\begin{figure}[t]
  \centering
   \begin{subfigure}[b]{0.4875\textwidth}
\includegraphics[width=\textwidth]{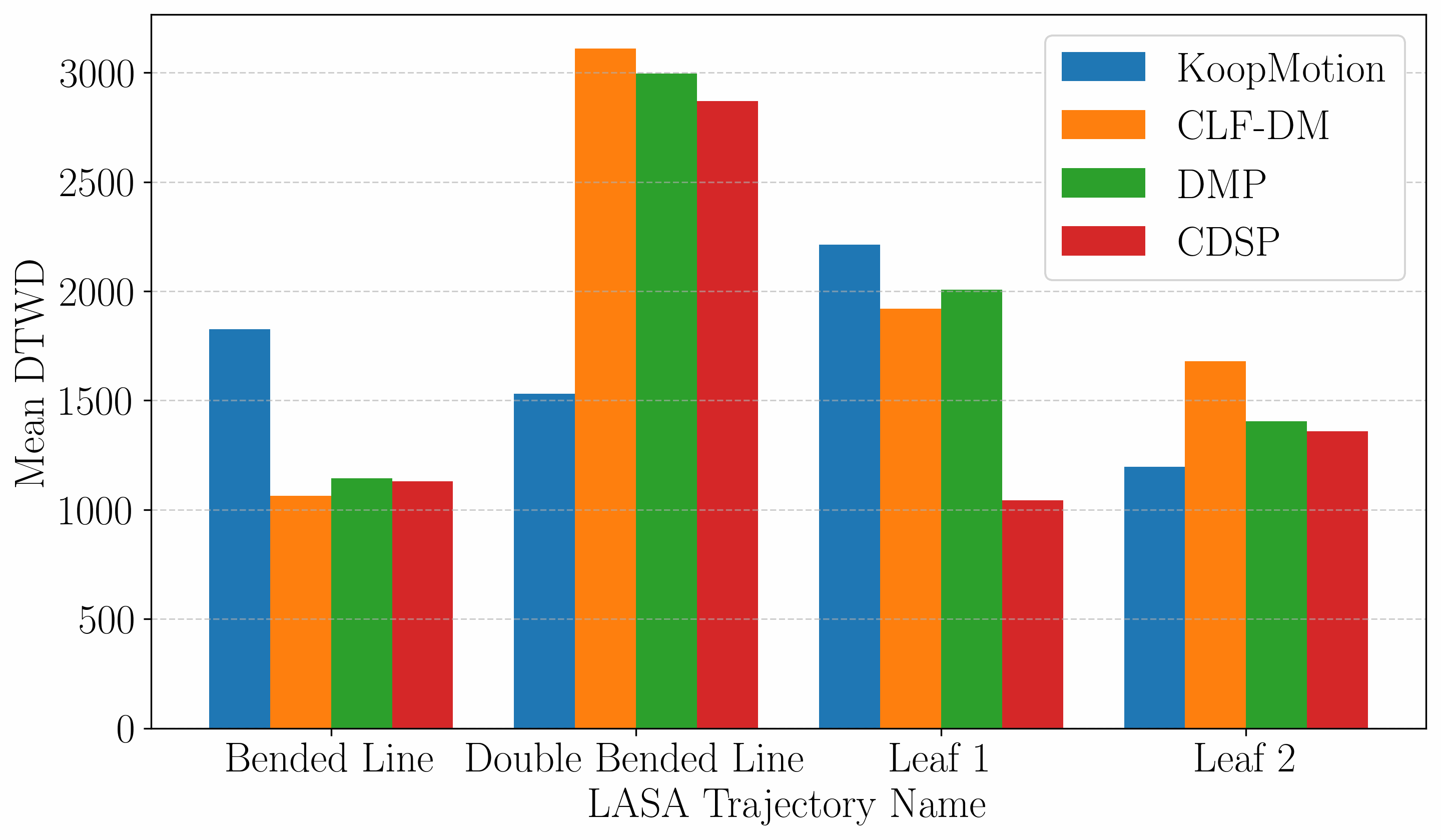}
    \end{subfigure}
       \begin{subfigure}[b]{0.4875\textwidth}
\includegraphics[width=\linewidth]{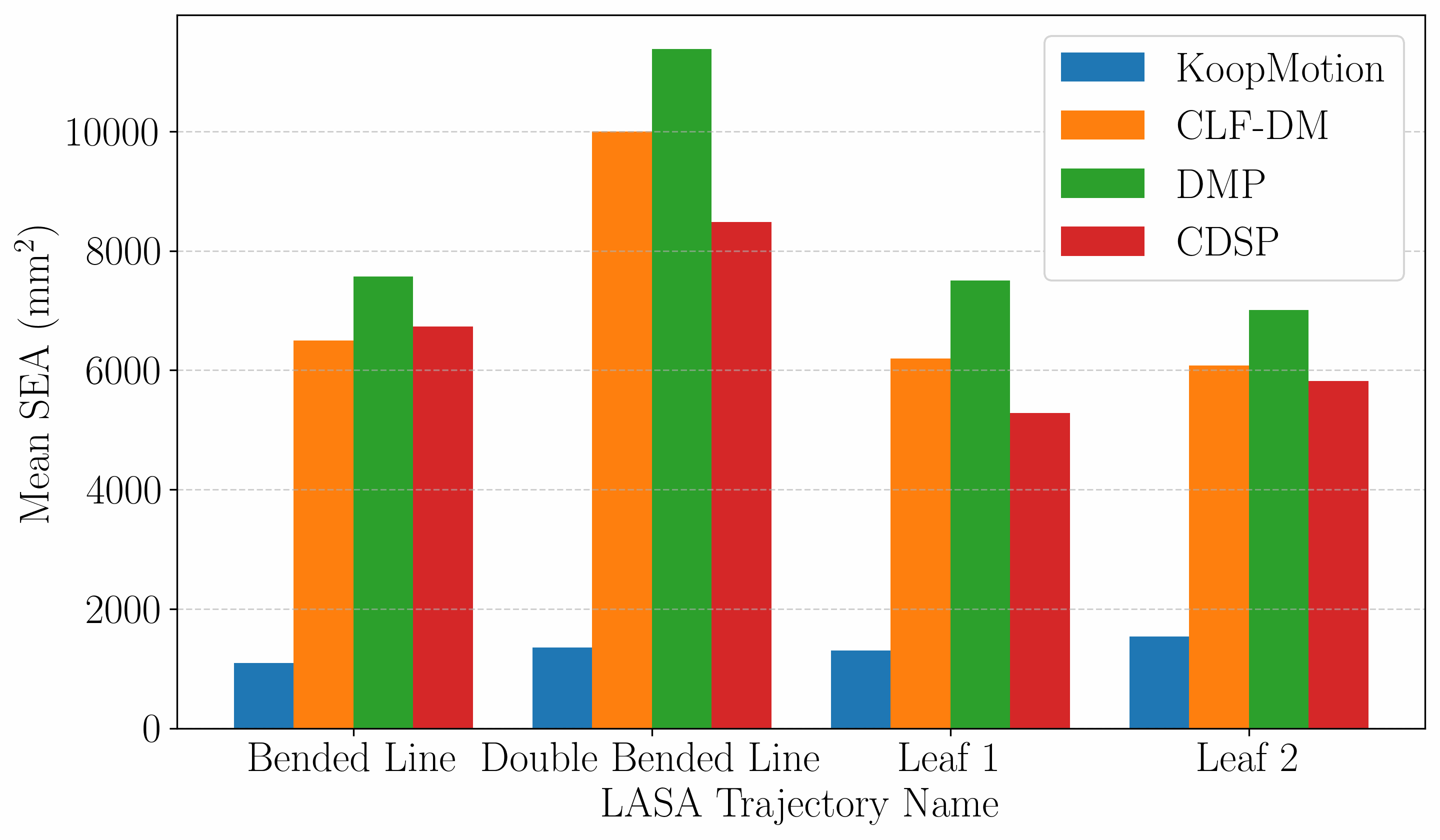}
    \end{subfigure}
    \caption{Performance evaluation of mean time dynamic time warping distance (DTWD) and mean swept error area (SEA) over 7 demonstrations for 4 nonlinear demonstrations of the LASA handwriting dataset. 
    For both metrics, the lower the value, the better. 
    Results have been recreated from Figure 3 found in \citep{ravichandar2017learning}. KoopMotion (ours) has overall comparable DTWD metrics compared to baselines.
    This can also be qualitatively observed when comparing the trajectories as shown in Fig. \ref{fig:lasa_demos}, and those in trajectories in \citep{ravichandar2017learning} and in \citep{khansari2014learning}. 
    On the other hand, KoopMotion significantly outperforms other baselines when comparing SEA metrics, which captures the spatiotemporal differences in the trajectories, and is difficult to visualize qualitatively otherwise.
    }
    \label{fig:metrics}
\end{figure}

\begin{figure}[t]
    \centering
    \includegraphics[width=1.0\textwidth]
        {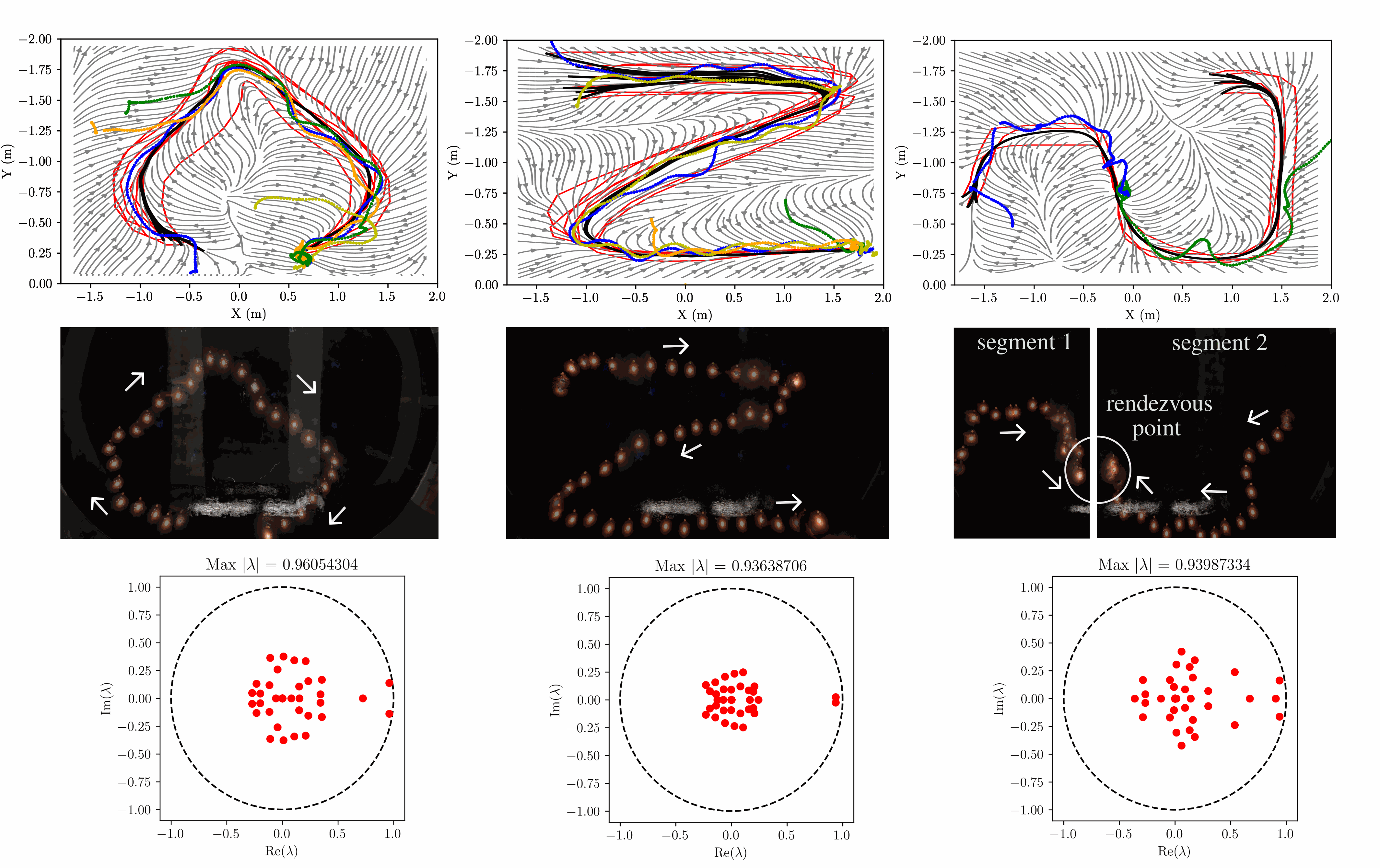}
        \caption{Experimental verification of KoopMotion with miniature autonomous surface vehicles operating in fluid flow tank on LASA \textsc{left} leaf (Leaf\_1), \textsc{center} letter `z' (ZShape), and \textsc{right} multi modal (Multi\_Models\_4) trajectory with two starting points. 
        \textsc{top row} shows training data motion demonstrations in red, learnt motions for same initial conditions as training data in black, attracting KoopMotion flow field in gray. 
        Other colors show experimental runs from different various conditions.
        \textsc{middle row} shows an example of snapshots of the autonomous surface vehicle in time, with convergence to the goal. 
        White arrows indicate direction of motion. 
        For the multi modal trajectories \textsc{right}, we show the trajectories of a robot starting from the left (segment 1), and starting from the right (segment 2), and separate the two real-world experiment trajectories to emphasize this. 
        This experiment demonstrates that the KoopMotion flow fields can be used for multi-robot rendezvous.
        We note that we did not fine-tune the existing velocity controller for these experiments.
        \textsc{bottom row} shows the spectral analysis of the system, showing the eigenvalues of the discrete-time Koopman operator.
        The magnitude of all eigenvalues are less than 1, showing system stability.}
        \label{fig:experiments_tank}
\end{figure}

\subsubsection{Baselines}
We compare our work to results shown in \citep{ravichandar2017learning}, including their CDSP algorithm, and the baselines therein: DMP \citep{ijspeert2013dynamical} and CLF-DM \citep{khansari2014learning}. 


\subsubsection{Metrics}
To quantitatively assess the performance of KoopMotion, we compute the dynamic time warping distance (DTWD) and swept error area (SEA), as defined in \citep{khansari2011learning} between each LASA demonstration and KoopMotion trajectories. 
DTWD is a measure of the similarity between the two temporal sequences that may vary in speed. 
This means that it measures a similarity in the shape of the demonstration and Koopman trajectories, without considering misalignment in time.
On the other hand, SEA penalizes both spatial and temporal misalignment between the two sets of trajectories. 
SEA is defined as:

\begin{equation}
    SEA_{mean} = \frac{1}{N}\Sigma_{n=1}^N \Sigma_{t=1}^T \mathcal{A} (\mathbf{x}^{demo}_t, \mathbf{x}^{demo}_{t+1}, \mathbf{x}^{Koop}_t, \mathbf{x}^{Koop}_{t+1})
\end{equation}
where $N$ is the number of demonstration trajectories, and $ \mathcal{A}(\mathbf{x}_1, \mathbf{x}_2, \mathbf{x}_3, \mathbf{x}_4) $ corresponds to the area between the points $\mathbf{x}_1$ to $\mathbf{x}_4$, and $\mathbf{x}^{demo}_t$ and $\mathbf{x}^{Koop}_t$ refers to the spatial location of the demonstration data and KoopMotion predicted data, at time $t$. 
The smaller the value of these metrics, the better the quality of the learnt dynamical system model.

Figure \ref{fig:metrics} summarizes the quantitative performance of KoopMotion in comparison to the baselines for $4$ types of motions. 
The results show the mean metrics over $7$ sets of demonstrations for each motion.
For both
KoopMotion (ours) has overall comparable DTWD metrics compared to baselines.
 This can also be qualitatively observed when comparing the trajectories as shown in Fig. \ref{fig:lasa_demos}, and those in trajectories in \citep{ravichandar2017learning} and in \citep{khansari2014learning}. 
On the other hand, KoopMotion significantly outperforms other baselines when comparing SEA metrics, which captures the spatiotemporal differences in the trajectories, and is difficult to visualize qualitatively otherwise.

\subsubsection{Evaluating System Spectral Properties}
Given that Koopman operator theory is a spectral method, we can apply linear tools and analysis on the learnt system.
This makes Koopman-based approaches more favorable over existing methods, including those that model dynamical systems with GMMs, NODEs, or Gaussian Process Regression methods, whose system stability cannot be analyzed as readily.
In this work, we learn a discrete time Koopman operator, defined by Equation \ref{eqn:koopman}.
Therefore, to assess the system stability, we can study the eigenvalues of the learnt $\hat{\mathcal{K}}$.
More specifically, we know that the system is asymptotically stable if and only if the magnitude of all of the eigenvalues of $\hat{\mathcal{K}}$ are less than 1: if $|\lambda_i|$ < 1 for all $i$, where $\lambda_i$ are eigenvalues of $\hat{\mathcal{K}}$.

The bottom row of Fig. \ref{fig:experiments_tank} includes the spectral analysis of the learnt Koopman operator that produced the gray flow fields estimated in the top row. 
Given that the magnitude of all of the eigenvalues of $\hat{\mathcal{K}}$,  $|\lambda_i|$ < 1, we know that these learnt dynamical systems are asymptotically stable. 


\subsection{Hardware Experiments}
We evaluate KoopMotion in 
an indoor laboratory experimental testbed, designed to enable experimental validation of motion control and coordination strategies.
This testbed is a $4.5 \text{m} \times 3.0 \text{m} \times 1.2 \text{m}$ water tank equipped with an OptiTrack motion capture system. 
A miniature autonomous surface vehicle (mASV) with differential drive, is used to evaluate the reference trajectories generated by KoopMotion.
Desired reference trajectories can be computed either by Euler integration of the learnt vector field, or through Koopman propagation. 
Both yield the same results.
To execute these trajectories on the mASV, an on-board velocity-controller is used.
We scale the learnt KoopMotion vector field such that the maximum vector magnitude does not exceed the maximum speed of the vehicle.
Then, as summarized in Fig. \ref{fig:koopmotion_overview}, the robot uses the KoopMotion vector field to iteratively determine its desired velocity, given its current position. 
This desired velocity $\mathbf{v}_{des}$ is used to compute the desired heading, $\theta = atan2 \Big( \frac{v_{{des}_y}}{v_{{des}_x}} \Big)$ and desired magnitude, $v_{des} = \|\mathbf{v}_{des}\|_2$.
These two parameters are sent to the velocity controller, which sends the required velocity commands to the left and right rear thrusters.

Figure \ref{fig:experiments_tank} shows the mASV operating in the tank.
The second row shows the mASV, depicted by its LEDs in time, using KoopMotion to track the desired reference trajectory.
The white arrows show the motion of the vehicle in the tank.
Given the smoothness of the learnt KoopMotion dynamical system, the mASV is able to follow the desired reference trajectory well.
Additionally, for the third case we have a successful demonstration of the mASV completing the multi-modal trajectory, as it starts from two different starting points, mimicking multi-robot rendezvous at the goal position.

In the first row, additional experimental runs are shown, showing that even if the robot starts in positions away from the demonstration motions (in colors other than red and black), it is still able to converge to the goal position (the end of the trajectory).
We note several additional details about these experiments.
First, as the robot is actuated and acts on the surrounding water, the motion of the fluid environment that the mASV operates in is non-static, possibly giving rise to the  wavy motion observed in the top row of Figure. \ref{fig:experiments_tank}.
Also, we did not fine-tune the PID controller for these experiments--other than scaling the vector field to the maximum velocity of the robot, this was a direct transfer of KoopMotion from sim2real.
We also note that the robot updates its velocity control using the current state.
However, the future predicted state or multiple predicted states could be used for smoother control.

\section{Conclusion}
In this work, we have introduced KoopMotion, a data-driven Koopman operator theoretic dynamics model that enables learning smooth motion plans that guide a system towards demonstration trajectories; along trajectories, and towards their goal positions.
Our experimental evaluations on the $2D$ LASA handwriting dataset and $3D$ Robocasa end-effector position dataset both qualitatively and quantitatively reveal the effectiveness of our novel divergence and convergence loss function terms.
Spectral analysis of the linear Koopman operator was also performed to assess the learnt system stability.
KoopMotion reference trajectories were additionally successfully evaluated on a miniature autonomous surface vehicle operating in a fluid flow tank. 
We show results trained on sparsely sampled datasets, highlighting the sample efficiency of our approach.
In future work, we are interested in performing evaluations on higher dimensional systems, namely for mobile aerial vehicles or manipulators, providing guarantees with methods such as those in \citep{guo2024surprising}, as well as extending work to capture greater variability in the demonstration motions.


\section{Limitations}
This work focuses on learning dynamical systems, learnt from demonstrations that provide reference trajectories that guide the system towards demonstration motions when away from the training data; guide the system along desired trajectories; and guide the system to convergence to a desired goal position.
While this work has empirically demonstrated the effectiveness of introducing loss terms that encourage effective flow fields for motion planning, we do not have guarantees for these system properties.
Moreover, the divergence loss, with the weighting selected in this work, causes the learnt trajectories to collapse onto one attractor.
This means that we can lose variation present in the motions, especially early on in the demonstration.
Additionally, while the method can provide reference trajectories, for deployment on real systems for motion planning, it relies on the availability of a low-level controller.
The method also does not account for the robot's dynamics when learning the dynamical system. 

\section{Acknowledgements}
We gratefully acknowledge the support of NSF DCSD-2121887.
\bibliography{bibliography}

\newpage 
\renewcommand{\thesection}{S\arabic{section}}
\setcounter{section}{0} 

\section*{Supplementary material for: KoopMotion: Learning Almost Divergence-Free Koopman Flow Fields
for Motion Planning}
\section{Additional Spectral Analyses---KoopMotion Eigenfunctions}\label{sec:supp_eigenfunctions}

In addition to performing stability analysis on the learnt system by studying the learnt Koopman operator, $\hat{\mathcal{K}}$, as shown in the main text, we can additionally visualize the system eigenfunctions.
Let $\Phi(\mathbf{y})$ denote the eigenfunctions of $\hat{\mathcal{K}}$ such  that the following is true, for a given state of interest $\mathbf{y}$:
\begin{equation}
    \mathcal{K}\Phi(\mathbf{y}) = \lambda \Phi(\mathbf{y})
\end{equation}

For a eigenvector $v_i$ of $\hat{\mathcal{K}}$, which provides a weighting of each of the learnt basis functions $\hat{\Psi}(\mathbf{y})$, the associated eigenfunction $\phi_i$, is defined by:

\begin{equation}
    \phi_i = v_i^T \Psi(\mathbf{y})
\end{equation}

Since we are interested in studying how the Koopman mode, as defined in \citep{mezic2013analysis}, affects or contributes to the vector field, we define $\mathbf{y} = \hat{\mathbf{x}}_{k+1} - \hat{\mathbf{x}}_{k}$.

\begin{figure}[h]
  \centering
\includegraphics[width=0.85\linewidth]{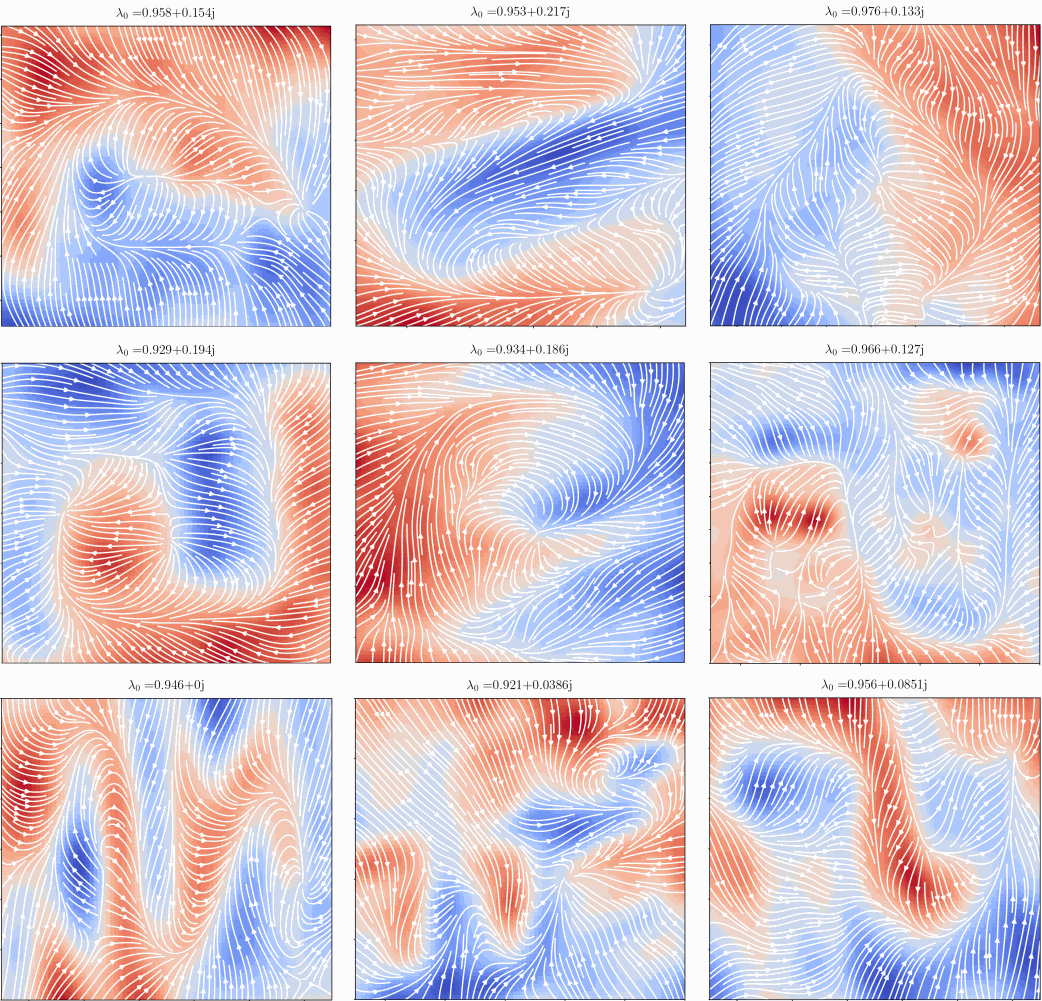}
  \caption{Eigenfunctions of KoopMotion models for shapes of the LASA dataset, based on the largest eigenvector of $\hat{\mathcal{K}}$. Eigenfunctions of the learnt Koopman operator represent groupings within the domain, where the space is partitioned into regions that exhibit similar dynamics.
  This is illustrated by partitions with the same color (eigenfunction value). 
  The shapes shown here from left to right, top to bottom are \textsc{BendedLine, ZShape, Leaf\_1, JShape\_2, PShape, Multi\_Models\_4, Sine, Multi\_Models\_3, Spoon}}
  \label{fig:supp_eigenfunctions}
\end{figure}

\section{KoopMotion Hyperparameters}
We train all shapes on the same hyperparameters to demonstrate the lack of reliance on defining a unique set of hyperparameters for an individual nonlinear shape 
This lack of reliance on fine-tuned hyperparameters contrasts prior work, which performance is dependent on a good choice of the number of Gaussian models that make up a GMM.

For this work, we use Adam, a stochastic gradient based optimizer, for optimization over the learnable parameters for the Fourier features $\hat{w}, \hat{b}$ and for the Koopman operator $\hat{\mathcal{K}}$, defined in Equation \ref{eqn:loss_function}. 
Based on the same Equation \ref{eqn:loss_function}, we select $\beta_k=1$, $\beta_d=0.01$, and $\beta_g=0.01$ for all examples shown.
We lift the system to a higher dimensional space of dimension $\nu = 1024$.
We train every KoopMotion model for $2200$ total iterations, with $200$ epochs and a batch size of $16$, with $168$ trajectory points.
We use a learning rate of $8e^{-4}$.
In practice, we use a low-rank form of the Koopman operator to minimize the number of parameters to be learnt. 
Meaning, we use $\hat{\mathcal{K}} = A B^T$, where $A \in \mathbf{R}^{(\nu + d)\times r}$ and $B \in \mathbf{R}^{(\nu + d)\times r}$ are matrices, and $\hat{\mathcal{K}} \in \mathbf{R}^{(\nu + d)\times (\nu + d)}$, where $\nu$ is the number of Fourier features, and $d$ represents the dimension of the original system, which for planar data such as the LASA dataset $d=2$, otherwise for 3d trajectories, $d=3$. 
We use $r=32$.
For a model with $\nu=1024$, the model is defined by approximately $70$k parameters.

\section{Modeling Trajectories in Higher Dimensional Spaces}\label{sec:modeling_3d_systems}
When trained on 3D position data, generated from the Robocasa simulator \citep{nasiriany2024robocasa}, we are successfully able to model the nonlinearities of the 3D trajectories in the training data. 
The eigenfunctions are able to additionally partition the dynamics of the three different modes of trajectories.
And finally, the learnt model is stable, as the magnitude of the largest eigenvalue is within the unit circle.

\begin{figure}[h]
  \centering
\includegraphics[width=0.975\linewidth]{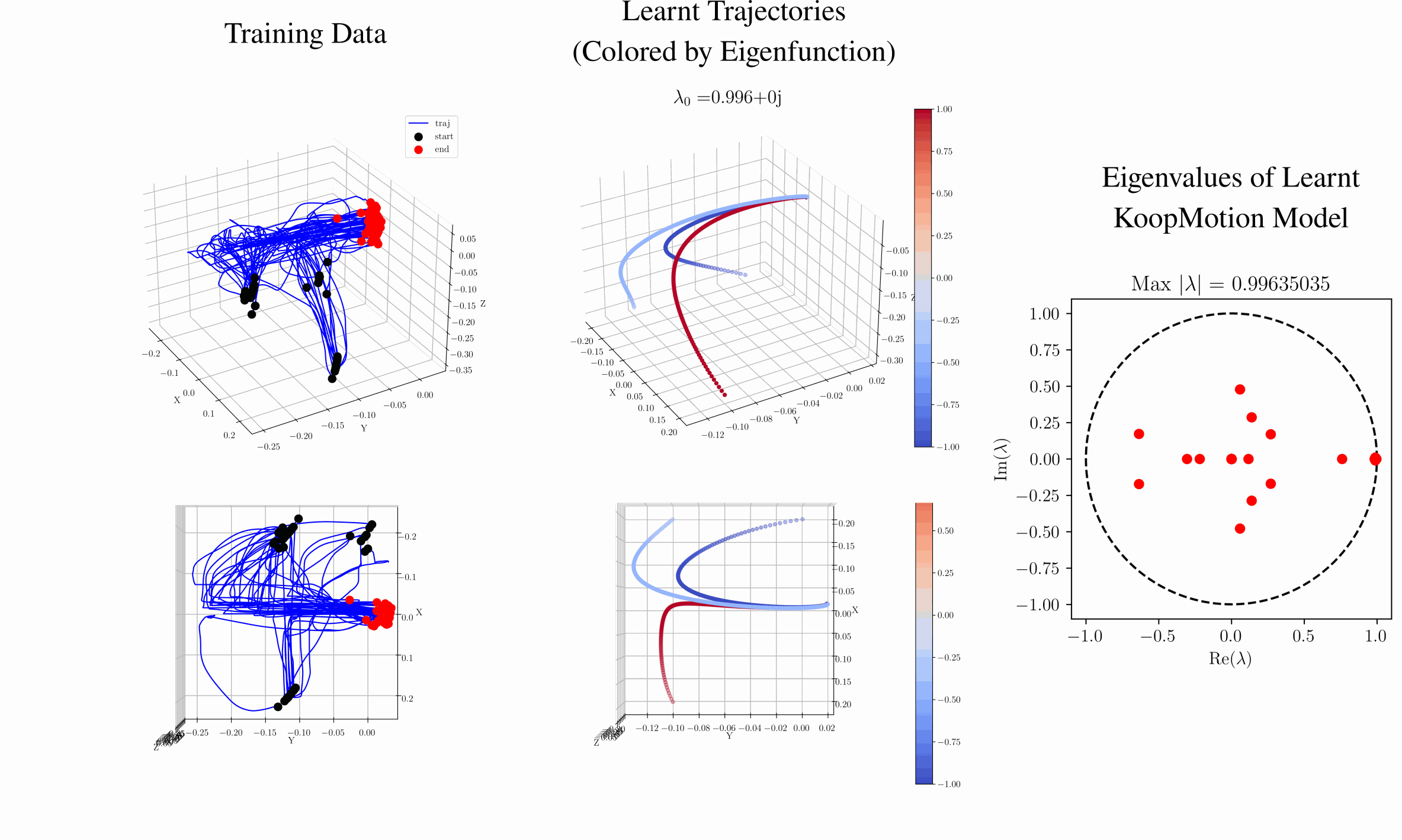}
  \caption{Learnt KoopMotion $\mathrm{R}^3$ dynamics for multi-modal trajectory data (3 separate groups of starting locations, converging to one goal position). \textsc{left} is the training data of an end effector operating in 3-dimensional space.
  \textsc{middle} is the learnt trajectories of points propagated by KoopMotion from initial conditions matching those in training data.
  These trajectories are colored by the eigenfunction value of these initial conditions.
  Given the three distinguishing colors, (dark blue, light blue, and red), similar to plots in Fig. \ref{fig:supp_eigenfunctions}, we demonstrate that eigenfunctions which are obtained from the learnt Koopman operator, partition the space into regions that exhibit similar dynamics.
  This partitioning is not readily available to the user with existing works, (e.g., GMMs, NODEs etc.).
  }
  \label{fig:supp_additional_experiments_3d}
\end{figure}

\section{Sparse Training Data}\label{sec:sparse_training_data}
As detailed in the main text, we train on 3\% of the original LASA dataset. 
To demonstrate how little data is used, we include examples below. 
Every demonstration has $25$ time-steps of the trajectory. 
We train on the $7$ demonstrations in the dataset. 
This gives rise to $168$ time-steps worth of trajectories in the training data ($24$ pairs of time-shifted data).

During evaluation using the metrics defined in the main text (DTWD and SEA), we compare time integrated data against the entire dataset, which has not been sub-sampled in time.

\begin{figure}[h]
  \centering
   \begin{subfigure}[b]{0.2\textwidth}
\includegraphics[width=1\linewidth]{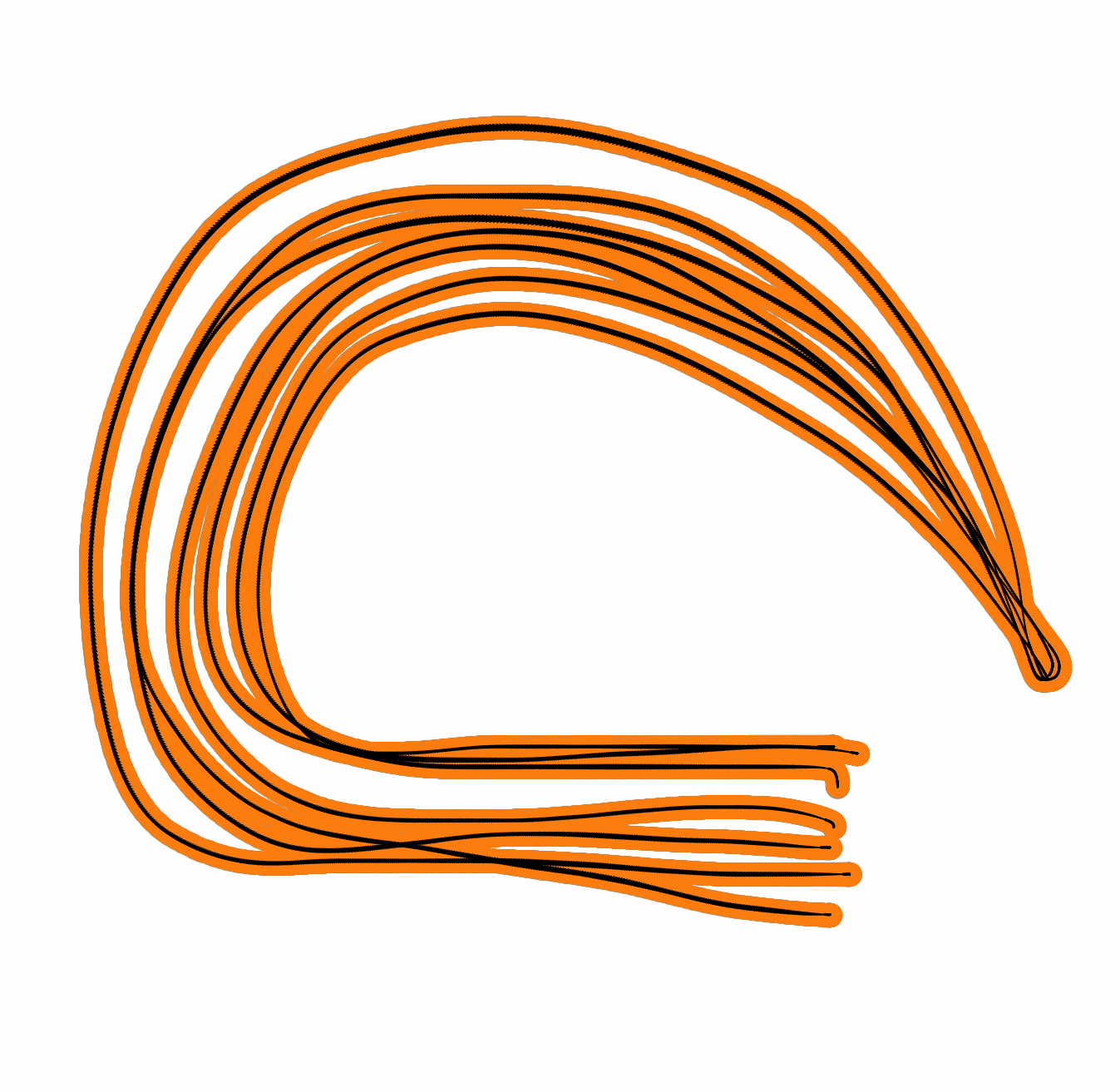}
  \caption{All training data.}
  \label{fig:training_data_all}
  \end{subfigure}
   \begin{subfigure}[b]{0.2\textwidth}
   \includegraphics[width=1\linewidth]{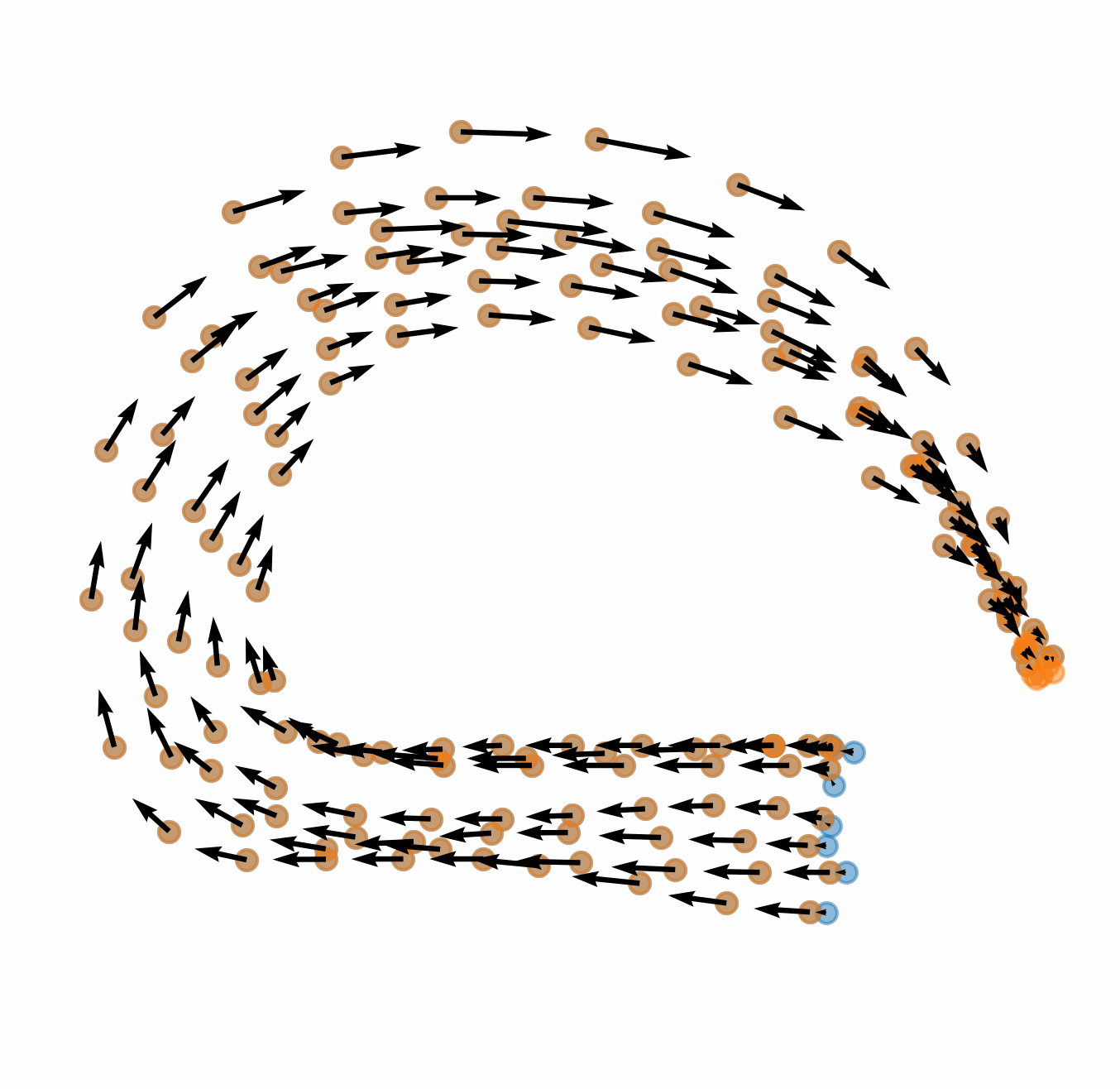}
  \caption{Training data used}
  \label{fig:sparse_data}
   \end{subfigure}
     \caption{Sparsity of the dataset.}
\end{figure}

\section{Additional Robot Experiments}
We include additional real-robot experiments that further validate the effectiveness of KoopMotion.
For these experiments, we limit the vehicle's motion to half of the maximum velocity, for smoother trajectory following using the KoopMotion vector fields, with a PID controller tuned for aggressive steering, as shown for following the \textsc{sine} shape well.

\begin{figure}[h]
  \centering
\includegraphics[width=0.85\linewidth]{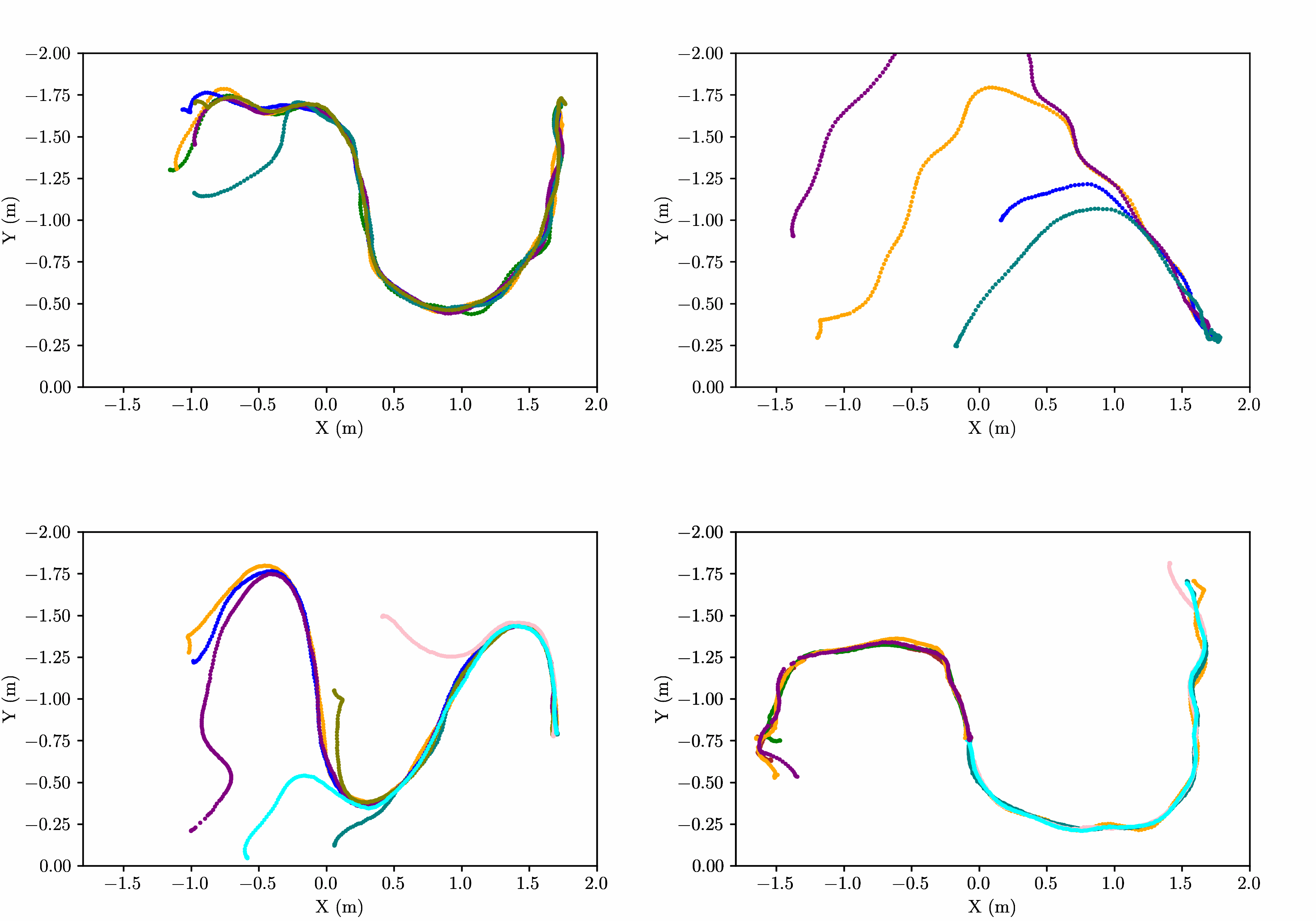}
  \caption{Additional results of the miniature autonomous surface vehicle using the vector fields for motion planning, where vector field is rescaled to half of the max vehicle velocity.
  The shapes shown here from left to right, top to bottom are \textsc{Spoon, Angle, Sine, BendedLine, ZShape, Multi\_Models\_4}}
  \label{fig:supp_additional_experiments}
\end{figure}

\section{Ablation Studies---Number of Fourier Features}
Across all of the shapes, we perform an ablation study on the effects of the number of Fourier features, $\nu$ on the two evaluation metrics defined in the main text, DTWD and SEA.

\begin{table}[h!]
\centering
\caption{Metric Average ($\pm$ Standard Deviation)}
\resizebox{\textwidth}{!}{%
\begin{tabular}{lccccc}
\toprule

Metric/$\nu$           & 500    & 750     & 1000     & 1250 & 1500     \\
\midrule
DTWD    & 1432.40 ($\pm$ 413.97)   & 1548.00 ($\pm$ 574.38 )   & 1595.49 ($\pm$ 537.12)   & 1576.91  ($\pm$ 504.71)   &1963.84($\pm$ 1380.01)    \\
SEA       & 148.35 ($\pm$ 69.55)   & 162.76 ($\pm$ 78.46)   & 169.39 ($\pm$ 73.95 )   & 174.94 ($\pm $68.15 )  &154.06($\pm$ 75.70 )   \\
\bottomrule
\end{tabular}}
\end{table}

\section{Ablation Studies---Loss Term Weighting}\label{sec:ablation_loss_term_weighting}

We ran an additional $1440$ experiments to understand the effects of varying the loss term weights. 
We perform pair-wise experiments for different sets of Koopman loss $\beta_k$, divergence loss $\beta_d$, and goal convergence loss $\beta_g$, where \textsc{experimented\_weights} = $\{0, 0.1, 1.0, 10.0\}$. 
By pair-wise, we mean that we keep one of the sets of weights constant, and vary the remaining two losses. 
This gives rise to $48$ experiments for each of the $30$ shapes of the LASA dataset, that allow us to understand the interactions between the loss terms. 
Note that before multiplying by the \textsc{experimented\_weights}, we inspect the gradient norms of each term during the initial iterations of training, and pre-adjust the weights such that the contributions of each of the terms to the total gradient are roughly comparable.
Here, we plot SEA (DWT results show similar trends).

\begin{figure}[h]
  \centering
\includegraphics[width=0.95\linewidth]{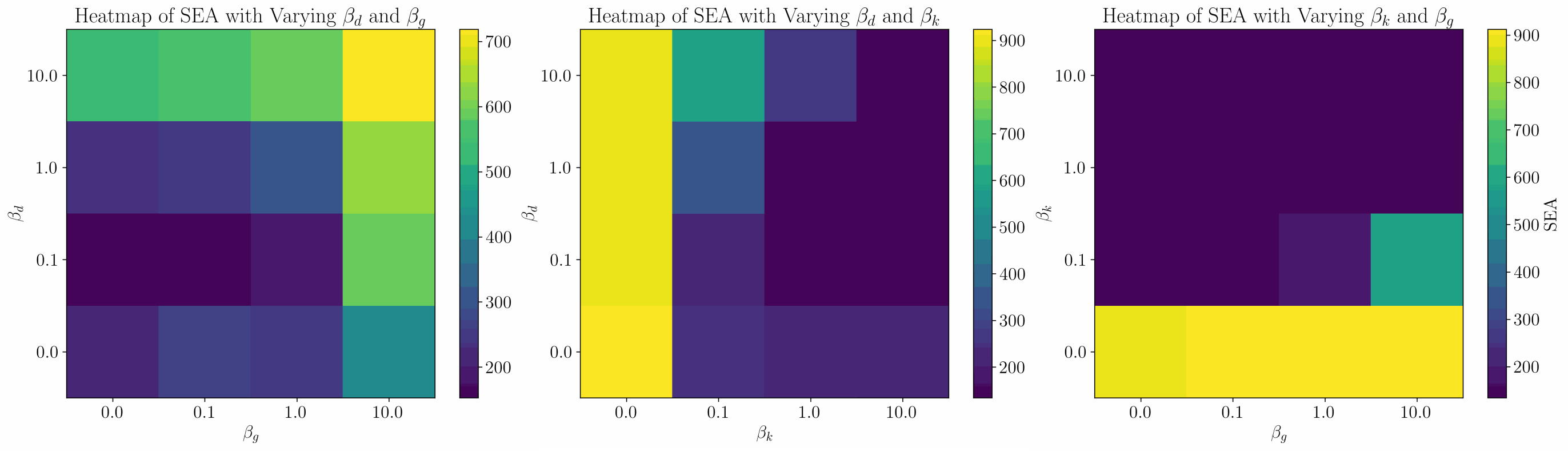}
  \caption{Ablation study to investigate the effects of varying the weights in the loss function. Note that the lower the SEA metric, the better the performance.}
  \label{fig:supp_additional_experiments_ablation_weights}
\end{figure}

\begin{figure}[h]
  \centering
\includegraphics[width=0.95\linewidth]{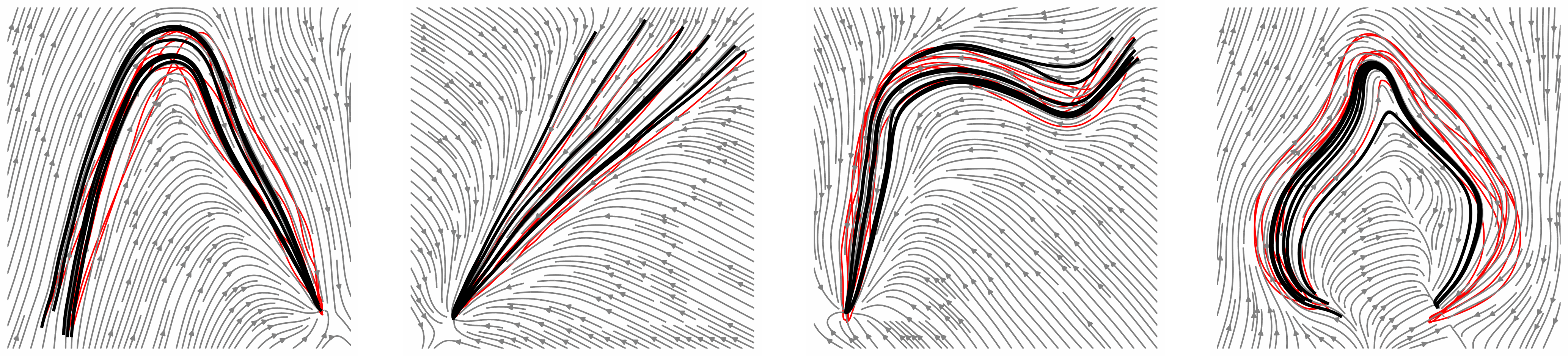}
  \caption{With the divergence loss term set to zero, we are able to achieve flows that do not collapse onto one another, but are less attracting for points away from the desired trajectory. These results demonstrate the flexibility of the approach, where loss weights can be adjusted based on the desired flow properties.}
\label{fig:supp_additional_experiments_non_collapsing}
\end{figure}




\end{document}